\documentclass{article}

    \PassOptionsToPackage{numbers, compress}{natbib}
    \usepackage[preprint]{neurips_2019}

\usepackage[utf8]{inputenc} % allow utf-8 input
\usepackage[T1]{fontenc}    % use 8-bit T1 fonts
\usepackage{url}            % simple URL typesetting
\usepackage{booktabs}       % professional-quality tables
\usepackage{amsfonts}       % blackboard math symbols
\usepackage{nicefrac}       % compact symbols for 1/2, etc.
\usepackage{microtype}      % microtypography

\usepackage{amsmath}
\usepackage{booktabs}
\usepackage{algorithm}
\usepackage{algorithmic}
\usepackage{times}
\usepackage{epsfig}
\usepackage{graphicx}
\usepackage{amsmath}
\usepackage{amssymb}
\usepackage{booktabs}       % professional-quality tables
\usepackage{amsfonts}       % blackboard math symbols
\usepackage{nicefrac}       % compact symbols for 1/2, etc.
\usepackage{microtype}      % microtypography
\usepackage{tabularx}
\usepackage{algorithm}
\usepackage{algorithmic}
\usepackage{epsfig}
\usepackage{epstopdf}
\usepackage{amsmath}
\usepackage{multirow}
\usepackage{subfigure}

\title{GmCN: Graph Mask Convolutional Network}

\author{
  Bo Jiang, Beibei Wang, Jin Tang and Bin Luo\\
  School of Computer Science and Technology, Anhui University\\
  \texttt{jiangbo@ahu.edu.cn} \\
}

\begin{document}

\maketitle

\begin{abstract}
Graph Convolutional Networks (GCNs) have shown very powerful for graph data representation and learning tasks.
%The graph convolution operation in GCNs can usually be regarded
%as a composition of feature aggregation/propagation and transformation.
Existing GCNs usually conduct feature
aggregation on a fixed neighborhood graph in which each node computes its representation by aggregating
the feature representations of all its neighbors which is biased by its own representation.
However, this fixed aggregation strategy is not guaranteed to be optimal for GCN based graph learning and also can be affected by some graph structure noises, such as incorrect or undesired edge connections.
To address these issues, we propose a novel Graph mask Convolutional Network (GmCN) in which
nodes can adaptively select the optimal neighbors in their feature aggregation to better serve GCN  learning.
GmCN can be theoretically interpreted by a regularization framework, based on which we derive a simple update algorithm to determine the optimal mask adaptively in GmCN training process.
Experiments on several datasets validate the effectiveness of GmCN.
\end{abstract}

\section{Introduction}

Recently, Graph Convolutional Networks (GCNs) have been commonly studied  to deal with graph representation and learning problems, such as semi-supervised learning, graph embedding, data clustering and so on~\cite{defferrard2016convolutional,kipf2016semi,velickovic2017graph,DGI,shchur2018pitfalls,zhou2018graph}.

One main aspect of GCNs is to develop a kind of specific and effective graph convolution operation in their layer-wise propagation. % on arbitrary structured graphs.
% The aim of GCNs is trying to define some reasonable convolution operations  on arbitrary structured graphs.
%These methods can generally be categorized into
%spatial methods and spectral methods. there is an increasing attention in trying to generalize CNNs to Graph Convolutional Networks (GCNs) to
%Spectral methods usually define graph convolution based on spectral representation of graphs.
% For example,
% Some recent works have also been proposed~\cite{velickovic2017graph}.
%they generally  define graph convolution operation based on spectral representation of graphs.
% For example,
Existing GCNs can generally be categorized into spectral and spatial methods.
Spectral methods aim to define graph convolution based on spectral representation of graphs.
For example, Bruna et al.~\cite{bruna2014spectral} define graph convolution based on the eigen-decomposition of graph Laplacian matrix.
Henaff et al.~\cite{henaff2015deep} define graph convolution by further introducing a spatially constrained spectral filter. %s to . % with smooth coefficients in order to make them spatially localized.
Defferrard et al.~\cite{defferrard2016convolutional} propose to define graph convolution by further approximating the spectral filters using Chebyshev expansion. % to avoid the high complexity. % of eigen-decomposition.
Kipf et al.~\cite{kipf2016semi} propose a simple Graph Convolutional Network (GCN) for semi-supervised learning by exploring the first-order approximation of spectral filters.
Different from spectral methods,
spatial methods focus on defining graph convolution by designing a specific aggregation function that aggregates the features of neighboring nodes.
For example, Li et al.~\cite{adaptive_GCN} present an adaptive graph CNNs. %, in which the graph is learned adaptively for GCN by employing a metric learning method.
Monti et al.~\cite{monti2017geometric} propose a unified mixture model CNNs (MoNet) for graph data  analysis. % and provide a unified generalization of CNN architectures on graphs.
Hamilton et al.~\cite{hamilton2017inductive} present  a general inductive representation and learning framework (GraphSAGE) for graph embedding.
%(eg, text attributes) to efficiently
Veli{\v{c}}kovi{\'c} et al.~\cite{velickovic2017graph} propose Graph Attention Networks (GAT). % that aggregates the features of neighboring nodes.
%Zhang et al.~\cite{CGNN} propose Capsule Graph Neural Network (CapsGNN) by adopting the concept of capsules to address the weakness in existing GNN-based graph embedding algorithms.
Petar et al.~\cite{DGI} propose Deep Graph Infomax (DGI) which provides a general unsupervised graph representation. % for learning node representation. % in unsupervised manner.
Jiang et al.,~\cite{jiang2019data} propose Graph Diffusion-Embedding Network (GEDNs) which proposes to integrate the general graph diffusion into graph convolution operation. % and e
They recently~\cite{jiang2019graph} also propose Graph Optimized Convolutional Network (GOCN) by
integrating graph construction and convolution together.
Klicpera et al.~\cite{klicpera2018predict} propose to combine GCN and PageRank together to derive an improved propagation scheme in layer-wise propagation.

In general, the graph convolution operation in the above GCNs can usually be regarded as a composition of feature aggregation/propagation and feature transformation.
Existing GCNs generally conduct feature aggregation on a fixed structured graph in which
each node obtains its representation recursively by aggregating/absorbing
the feature representations of its all neighbors. % (biased by its own representation).
However,  this fixed `full' aggregation strategy may be not guaranteed to be optimal for GCN based graph learning tasks. %, such as semi-supervised classification.
Also, it can be affected by the graph structure noises, such as incorrect or undesired edge connections.

To overcome these issues, we propose Graph mask Convolutional Network (GmCN) for graph data representation and learning.
The main idea behind GmCN is that it learns how to aggregate feature information \textbf{selectively} from a node’s local neighborhood set.
Specifically, in graph mask convolution (GmC) operation, each node can adaptively select the desirable/optimal neighbors from its full neighborhood set to conduct feature aggregation, which allows GmC to better serve the final graph  learning tasks. %, such as graph node classification.
GmC is achieved by employing  an adaptive and learnable mask guided aggregation mechanism. % to design a novel graph mask convolution operation.
More importantly, to determine the optimal selection mask in GmC, we derive a new unified regularization framework for GmC interpretation and thus develop a simple update algorithm to learn the optimal mask adaptively in GmCN training process.
%
%The main advantage of GmCN is that the learned representation of data can provide useful ``weakly" supervised information for
%learning a better graph which simultaneously facilitates graph convolutional representation and learning.
%%
%
Overall, the main contributions of this paper are summarized as follows:
\begin{itemize}
  \item We propose a novel graph mask convolution (GmC) operation which simultaneously learns the topological structure of each node’s neighborhood for feature aggregation as well as the linear (or nonlinear) transformation for feature representation.
  \item We develop a unified regularization framework for GmC interpretation and optimization.
  \item Based on the proposed GmC, we propose  Graph mask Convolutional Network (GmCN) for
    graph node representation and semi-supervised learning.
\end{itemize}
Experiments validate the effectiveness of GmCN on semi-supervised learning tasks.

%One important aspect of GCNs is the graph structure representation of data. In general, the graph data we feed to existing GCNs should be a single graph which is obtained
%from either domain knowledge ({e.g.}, social network) or human establishment, such as $k$-NN graph.
%However, there are three main issues.
%First, traditional human established graphs generally use fixed parameters to determine the graph structure and thus are usually sensitive to the local noises and errors.
%Second, the graphs obtained from domain knowledge or established by human are generally independent of graph convolutional learning, which thus are not guaranteed to be optimal for graph convolutional representation/learning in GCNs.
%Third, existing GCNs are generally hard to deal with multiple graphs, although some heuristic fusion strategies can be utilized for multiple graph convolutional network learning~\cite{simonovsky2017dynamic,schlichtkrull2018modeling,zhuang2018dual}.
%It is known that the convolution operation on multiple graphs is not as well-defined as on single graph.

%outperforms the state-of-the-art graph neural network methods.

\section{Revisiting GCN}
%
%Recently, Graph Convolutional Networks (GCNs) have been widely studied for graph data representation and learning~\cite{defferrard2016convolutional,kipf2016semi,henaff2015deep,velickovic2017graph}.
%The core aspect of GCNs is the specific definition of graph convolution operation in GCN layer-wise propagation.
In this section, we briefly review the widely used GCN model proposed in work~\cite{kipf2016semi}.
Given a set of node features $\textbf{H}=(\textbf{h}_1,\textbf{h}_2\cdots \textbf{h}_n) \in \mathbb{R}^{n\times d}$ and graph structure $\textbf{A}\in \mathbb{R}^{n\times n}$, GCN~\cite{kipf2016semi} defines a kind of graph convolution as, %conducts the layer-wise propagation as, % defines the graph attention
%%
%\begin{align}\label{EQ:layer00}
%{\textbf{H}'}&= \sigma\big((\textbf{I}+\textbf{D}^{-\frac{1}{2}}\textbf{A}\textbf{D}^{-\frac{1}{2}})\textbf{H}\Theta\big) \\
%&=\sigma\big((\textbf{I}+\hat{\textbf{A}})\textbf{H}\Theta\big),  \,\,\, \mathrm{where} \,\,\, \hat{\textbf{A}} = \textbf{D}^{-\frac{1}{2}}\textbf{A}\textbf{D}^{-\frac{1}{2}}
%%&\textbf{H}^{(l+1)} = \sigma\big(\widetilde{\textbf{H}}^{(l)}  {\Theta}^{(l)}\big)
%\end{align}
%%
%
\begin{align}\label{EQ:layer00}
{\textbf{H}'}&= (\textbf{D}^{-\frac{1}{2}}\textbf{A}\textbf{D}^{-\frac{1}{2}}+\textbf{I})\textbf{H}\Theta %\\
%&=\sigma\big((\textbf{I}+\hat{\textbf{A}})\textbf{H}\Theta\big),  \,\,\, \mathrm{where} \,\,\, \hat{\textbf{A}} = \textbf{D}^{-\frac{1}{2}}\textbf{A}\textbf{D}^{-\frac{1}{2}}
%&\textbf{H}^{(l+1)} = \sigma\big(\widetilde{\textbf{H}}^{(l)}  {\Theta}^{(l)}\big)
\end{align}
where $\textbf{D}$ is a diagonal matrix with elements $\textbf{D}_{ii}=\sum_j\textbf{A}_{ij}$ and ${\Theta}\in \mathbb{R}^{d\times \tilde{d}}$ is a layer-specific trainable weight matrix. %, and $\sigma(\cdot)$ denotes an activation function, such as $\mathrm{ReLU}(\cdot) = \max(0,\cdot)$.
Intuitively, the graph convolution operation Eq.(\ref{EQ:layer00}) can be decomposed into two operations, i.e.,
%%
%\begin{align}\label{EQ:layer1}
%\mathrm{S1:}\,\,\, {\textbf{U}} = \hat{\textbf{A}}\textbf{H} + \textbf{H},\,\,\, \,\,\,\,\,\, \mathrm{S2:} \,\,\,  {\textbf{H}'}  = \sigma({\textbf{U}}{\Theta})
%\end{align}
%%
%
\begin{align}\label{EQ:layer1}
\mathrm{S1:}\,\,\,\,\, {\textbf{U}} = \hat{\textbf{A}}\textbf{H} + \textbf{H},\,\,\, \,\,\,\,\,\, \mathrm{S2:} \,\,\,\,\,  {\textbf{H}'}  = {\textbf{U}}{\Theta}
\end{align}
where $\hat{\textbf{A}} = \textbf{D}^{-\frac{1}{2}}\textbf{A}\textbf{D}^{-\frac{1}{2}}$.
 The operation S1 provides a kind of feature aggregation on normalized graph $\hat{\textbf{A}}$ while operation S2 presents a transformation for node features. Recent works~\cite{klicpera2018predict,li2018dcrnn_traffic,jiang2019data} also suggest to employ a more general $T$-step truncated random walk model to conduct feature aggregation in step S1 as
\begin{align}\label{EQ:layer2}
 {\textbf{U}}^{(t)} = \alpha\hat{\textbf{A}}{\textbf{U}}^{(t-1)} + (1-\alpha)\textbf{H}
 \end{align}
where $t=1,2\cdots T$ and ${\textbf{U}}^{(0)} =\textbf{H} $, and parameter $\alpha\in(0,1)$ balances two terms. When $t=1$, Eq.(\ref{EQ:layer2}) degenerates to Eq.(\ref{EQ:layer1}) by ignoring the weighting parameter $\alpha$.

In each hidden layer of GCN, an activation function $\sigma(\cdot)$ is further conducted on $\textbf{H}'$ to obtain nonlinear representations.
 % used to nonlinearity of feature transformation. denotes , such as $\mathrm{ReLU}(\cdot) = \max(0,\cdot)$.
%The last layer of GCN outputs the final representations of graph nodes, which can be used for many learning tasks, such as clustering, visualization and node classification etc.
%Here, we focus on semi-supervised classification.
%In this paper, we focus on semi-supervised classification.
For node classification task, a softmax activation function is further used in the last layer to output the label predictions for graph nodes. %, defined as
%% $\textrm{softmax}(\textbf{x}) = \frac{\exp(\textbf{x}_i)}{\sum_i \exp(\textbf{x}_i)}$,
% is further conducted on each row of the final output feature map matrix $\textbf{H}^{(K)}$.
%Let $\textbf{Z}=\textrm{softmax}(\textbf{H}^{(K)} )\in \mathbb{R}^{n\times c}$ be the final output, where $c$ denotes the number of class. Then $\textbf{Z}$ provides a kind of label prediction for graph nodes.
The weight matrix $\Theta$ of  each hidden layer % $\{{\Theta}^{(0)},{\Theta}^{(1)},\cdots {\Theta}^{(K-1)}\}$
is optimized by minimizing the cross-entropy loss defined on labelled nodes~\cite{kipf2016semi}.
%%
% \begin{equation}
%\mathcal{L}_{\textrm{Semi-GCN}} = -\sum\nolimits_{i\in L} \sum^c\nolimits_{j=1} {Y}_{ij}\mathrm{ln} {P}_{ij}
% \end{equation}
%%
%where ${L}$ indicates the set of labeled nodes and each row ${Y}_{i\cdot}, i\in L$ of ${Y}$ denotes the corresponding label indication vector for the $i$-th labeled node.

\section{Graph-mask Convolutional Network}

In this section, we present our Graph mask Convolutional Network (GmCN) model.
Our GmCN is motivated by the aggregation operation Eq.(\ref{EQ:layer2}) or Eq.(\ref{EQ:layer1}), which is further analyzed as follows.
%we rewrite update Eq.(\ref{EQ:layer00}) as
%%
%\begin{align}\label{EQ:gcn_diffusion}
%\textbf{H} = (\textbf{I}+\hat{\textbf{A}})\textbf{H}
%\end{align}
%%
Let ${\textbf{U}}^{(t)}=({\textbf{u}}^{(t)}_1,{\textbf{u}}^{(t)}_2\cdots {\textbf{u}}^{(t)}_n)$ and $\textbf{H}=(\textbf{h}_1,\textbf{h}_2\cdots \textbf{h}_n)$,
then Eq.(\ref{EQ:layer2}) is formulated  as %Let $\textbf{H}=(\textbf{H})$That is,
\begin{align}\label{EQ:layer22}
{\textbf{u}}^{(t)}_i = \alpha\sum\nolimits_{j\in \mathcal{N}_i}\hat{\textbf{A}}_{ij}{\textbf{u}}^{(t-1)}_j  +(1-\alpha)\textbf{h}_i
\end{align}
where $\mathcal{N}_i$ denotes the neighborhood set of the $i$-th node.
% where $\hat{\textbf{A}} = \textbf{D}^{-\frac{1}{2}}\textbf{A}\textbf{D}^{-\frac{1}{2}}$.
From Eq.(\ref{EQ:layer22}), we can note that % GCN employs an one-step feature aggregation on normalized graph $\hat{\textbf{A}}$ (biased by feature itself) to obtain contextual feature representation in layer-wise propagation.
 in GCN layer-wise propagation, each node aggregates the feature information from its \emph{all neighboring} nodes on normalized graph $\hat{\textbf{A}}$ (biased by its own feature $\textbf{h}_i$) to obtain the context-aware representation, and $\alpha\in(0,1)$ denotes the fraction of feature information that the $i$-th node  receives from its neighbors.

However, first, this fixed `full' aggregation strategy may be not guaranteed to be optimal for GCN based specific learning tasks, such as semi-supervised classification.
Second, this `full' aggregation can be affected by the graph structure noises, such as incorrect or undesired edge connections, disturbance of edge weights and so on.
To overcome these issues,
% However, this `full' aggregation may be insensitive to graph noises (undesired or incorrect edge connections).
we propose to adaptively select some desired neighboring nodes from its neighborhood set $\mathcal{N}_i$ to conduct feature aggregation.
%\emph{Obviously, this one-step diffusion does not return  the equilibrium representation of feature
%diffusion which thus may lead to weak contextual feature representation.}
This can be achieved by using the mask weighted aggregation model as %general diffusion-embedding network.
\begin{align}\label{EQ:layer3}
{\textbf{u}}^{(t)}_i = \alpha\sum\nolimits_{j\in \mathcal{N}_i}\textbf{M}_{ij}\hat{\textbf{A}}_{ij}{\textbf{u}}^{(t-1)}_j  + (1-\alpha)\textbf{h}_i
\end{align}
where $\textbf{M}\in \mathbb{R}^{n\times n}, \textbf{M}_{ij}\in \{0,1\}$ denotes the mask matrix in which $\textbf{M}_{ij}=1$ indicates that the $j$-th node in $\mathcal{N}_i$ is selected and $\textbf{M}_{ij}=0$ otherwise. This mask matrix is learned adaptively, as discussed in Section 4 in detail. %, and $\alpha\in(0,1)$ denotes the fraction of feature information that the $i$-th node  receives from its neighbors. % on normalized graph $\hat{{A}}$.
Based on this mask weighted aggregation, we then propose our \emph{Graph mask Convolution} (GmC) operation as, % the layer-wise propagation in GmCN as
\begin{align}\label{EQ:layer4}
 & {\textbf{U}}^{(t)} = \alpha(\textbf{M}\odot\hat{\textbf{A}}){\textbf{U}}^{(t-1)} + (1-\alpha)\textbf{H}, \,\,\, t=1,2\cdots T \\
& {\textbf{H}'}  = {\textbf{U}}^{(T)}{\Theta}
\end{align}
where ${\textbf{U}}^{(0)} = \textbf{H}$ and $\odot$  represents element-wise multiplication. Parameters  $\{\textbf{M},{\Theta}\}$
denote the layer-specific trainable mask matrix and weight matrix, respectively.
Similar to GCN~\cite{kipf2016semi}, in layer-wise propagation, an activation function $\sigma(\cdot)$ is further conducted on $\textbf{H}'$ to obtain nonlinear representations.
%
%is a layer-specific trainable weight matrix and $\textbf{M} \in \mathbb{R}^{n\times n}, \textbf{M}_{ij}\in \{0,1\}$ denotes the layer-specific  mask matrix needing to be optimized. $\sigma(\cdot)$ denotes an activation function, such as $\mathrm{ReLU}(\cdot) = \max(0,\cdot)$.
% In the following, we present an effective algorithm to optimize the mask matrix $\textbf{M}$.
Note that, when $T=1$, our GmC has a simple formulation as
\begin{align}\label{EQ:layer11}
 {\textbf{H}'}  = \big(\alpha\textbf{M}\odot\hat{\textbf{A}} + (1-\alpha)\textbf{I}\big)\textbf{H} {\Theta}
\end{align}

\textbf{Remark}.
In contrast to GCN which only has weight parameter $\Theta$ in each layer to conduct feature transformation, GmCN has two kinds of parameters $\{\textbf{M}, \Theta\}$ in which $\textbf{M}$ is employed to
select the desired neighbors for feature aggregation while $\Theta$ is used to conduct feature transformation.
We will present the detail comparisons between GmCN and other related GCN works in Section 5.
In the following, we first propose an effective algorithm to learn $\textbf{M}$. We summarize the complete architecture and parameter training of our GmCN network in Section 5.

%In addition, in this paper, we explore a more flexible $t$-step aggregation as
%%
%\begin{align}\label{EQ:propagation_rwr}
%\widetilde{\textbf{H}}^{(t+1)}= \alpha \hat{{A}} \widetilde{\textbf{H}}^{(t)} + (1-\alpha) \widetilde{\textbf{H}}\textstyle =  \big[(\alpha \hat{{A}})^{t}+(1-\alpha)\sum^{t}_{i=0}(\alpha \hat{{A}})^i\big] \widetilde{\textbf{H}}
%%&\textstyle =  \big[(\alpha \hat{{A}})^{t+1}+(1-\alpha)\sum^{t}_{i=0}(\alpha \hat{{A}})^i\big] H
%\end{align}
%%

%
%%%
%%
%\begin{align}\label{EQ:layer00}
%&\widetilde{\textbf{H}}^{(l)} = (\textbf{I}+\textbf{D}^{-\frac{1}{2}}\textbf{A}\textbf{D}^{-\frac{1}{2}})\textbf{H}^{(l)} \\
%&\textbf{H}^{(l+1)} = \sigma\big(\widetilde{\textbf{H}}^{(l)}  {\Theta}^{(l)}\big)
%\end{align}
%%
%where $l=0,1,\cdots L-1$ and  $\textbf{D}$ is a diagonal degree matrix with elements $\textbf{D}_{ii}=\sum_j\textbf{A}_{ij}$. Parameters ${\Theta}^l\in \mathbb{R}^{d_l\times {d}_{l+1}}$ is a layer-specific trainable weight matrix, and $\sigma(\cdot)$ denotes an activation function, such as $\mathrm{ReLU}(\cdot) = \max(0,\cdot)$.

\section{Regularization Framework and Mask Optimization}

\subsection{Regularization framework}

% In this section, we present our Graph Optimized Convolutional Network (GOCN) model.
Our mask optimization is motivated based on re-interpretation of feature aggregation in GCN by using a regularization framework. %~\cite{jiang2019data,jiang2019graph}.
Formally,
%In the following, we first present our regularization reformulation of graph convolution in \S 3.1.
%Based on it, we then derive a  unified framework of graph optimization-convolution operation in \S 3.2.
%Finally, we present our GOCN architecture in \S 3.3.
 we can  show that, Eq.(3) provides a $T$-step power iteration solution to the following regularization framework~\cite{jiang2019data,jiang2019graph,zhou2004learning}\footnote{When $T$ is large enough, the update Eq.(3) converges to the optimal solution as $\textbf{U}^*=
 (\textbf{I}-\alpha\hat{\textbf{A}})\textbf{H}$. It is identical to the optimal solution of this problem Eq.(9) given as $\textbf{U}^*=(1-\alpha)
 (\textbf{I}-\alpha\hat{\textbf{A}})\textbf{H}$ by ignoring the scaling coefficient, as proved in work~\cite{zhou2004learning}.},
\begin{equation}\label{EQ:propagation_opt0}
\min_{\textbf{U}} \,\, \mathcal{R}_{\mathrm{GCN}}(\textbf{U}) =\mathrm{Tr} [\textbf{U}^{\mathrm{T}}(\textbf{I} - \hat{\textbf{A}}) \textbf{U}] + \mu \|\textbf{U} - \textbf{H}\|^2_F
\end{equation}
where $\mu = \frac{1}{\alpha} - 1$ is a replacement parameter of $\alpha$. $\mathrm{Tr}(\cdot)$  denotes the trace function and $\|\cdot\|_F$ denotes the Forbnuous norm of matrix.
 % This is because the optimal $\textbf{U}$ can be obtained by using the following power iteration algorithm.
%%
%\begin{equation}\label{EQ:iteration}
%\textbf{U}^{(t+1)} = \alpha\hat{\textbf{A}} \textbf{U}^{(t)} + (1-\alpha)\textbf{H}
%\end{equation}
%%
%where $t=0,1\cdots$ and $\textbf{U}^{(0)}=\textbf{H}$.
Using this representation, we can thus derive a regularization framework for our GmCN (Eq.(6) or Eq.(8)) as
\begin{align}\label{EQ:propagation_opt1}
&\min_{\textbf{M},\textbf{U}} \,\, \mathcal{R}_{\mathrm{GmCN}}(\textbf{M},\textbf{U}) =  \mathrm{Tr} [\textbf{U}^{\mathrm{T}}(\textbf{I} - \textbf{M}\odot\hat{\textbf{A}}) \textbf{U}] + \mu \|\textbf{U} - \textbf{H}\|^2_F + \gamma\|\textbf{M}\|^2_F\\
&\, s.t. \,\,\,\, \textbf{M}_{ij}\in\{0,1\}
\end{align}
where $\gamma >0$ and the last term is used to control the sparsity (number of non-zeros) of mask matrix $\textbf{M}$, i.e., larger parameter $\gamma$ leads to more sparse $\textbf{M}$.
From optimization aspect, the discrete constraints $\textbf{M}_{ij}\in\{0,1\}$ make the problem be combinational and thus hard to be optimized.
Therefore, we first relax the discrete constraints to continuous ones and solve
\begin{align}\label{EQ:propagation_opt}
&\min_{\textbf{M},\textbf{U}} \, \mathcal{R}_{\mathrm{GmCN}}(\textbf{M},\textbf{U}) = \mathrm{Tr} [\textbf{U}^{\mathrm{T}}(\textbf{I} - \textbf{M}\odot\hat{\textbf{A}}) \textbf{U}] + \mu \|\textbf{U} - \textbf{H}\|^2_F + \gamma\|\textbf{M}\|^2_F \\
&\, s.t. \,\,\,\,  \textbf{M}\textbf{1}^{\mathrm{T}} = \textbf{1}^{\mathrm{T}}, \textbf{M}\geq 0% \{0,1\}
\end{align}
where $\textbf{1}=(1,1\cdots 1)$.
Then, we utilize a post-discretization step to obtain the final discrete mask $\textbf{M}$, as summarized in Algorithm 1.
% In the following, we derive an effective algorithm to solve Eq.(\ref{EQ:propagation_opt}).

\subsection{Mask optimization}

The optimal $\textbf{M}$ and $\textbf{U}$ can be obtained  alternatively conducting the following \textbf{Step 1} and \textbf{Step 2} until convergence.

\textbf{Step 1}. Solving $\textbf{M}$ while fixing $\textbf{U}$, the problem becomes
\begin{align}\label{EQ:opt_M}
\min_{\textbf{M}} \,\, \mathrm{Tr} [\textbf{U}^{\mathrm{T}}(\textbf{I} - \textbf{M}\odot\hat{\textbf{A}}) \textbf{U}] + \gamma\|\textbf{M}\|^2_F\,\,\,\,\,\, s.t. \,\,\,\,  \textbf{M}\textbf{1}^{\mathrm{T}} = \textbf{1}^{\mathrm{T}}, \textbf{M}\geq 0% \{0,1\}
\end{align}
It is rewritten as
\begin{align}\label{EQ:opt_M1}
\min_{\textbf{M}} \,\, \sum_{i,j}\nolimits -(\textbf{M}\odot\hat{\textbf{A}})_{ij}(\textbf{U}\textbf{U}^{\mathrm{T}})_{ij} + \gamma \|\textbf{M}\|^2_F\,\,\,\,\,\,\, s.t. \,\,\,\,  \textbf{M}\textbf{1}^{\mathrm{T}} = \textbf{1}^{\mathrm{T}}, \textbf{M}\geq 0% \{0,1\}
\end{align}
Let $\textbf{Z}=\hat{\textbf{A}}\odot (\textbf{U}\textbf{U}^{\mathrm{T}})$, then Eq.(\ref{EQ:opt_M1}) is equivalent to
\begin{align}\label{EQ:opt_M2}
\min_{\textbf{M}}  \,\,  \|\textbf{M} - \frac{1}{2\gamma} \textbf{Z} \|^2_F \,\,\,\,\,\,\, s.t. \,\,\,\,  \textbf{M}\textbf{1}^{\mathrm{T}} = \textbf{1}^{\mathrm{T}}, \textbf{M}\geq 0% \{0,1\}
\end{align}
This is a constrained projection problem which can be
solved efficiently via a successive projection algorithm, as discussed in work~\cite{zass2007doubly}.
The algorithm is summarized in Algorithm 1. Note that, in practical, we use a $T$-step truncated iteration to obtain an approximate solution. We will show this  in Algorithm 2 in detail. % for the trade-off of effectiveness and efficiency.

\begin{algorithm}[h]
\caption{Mask optimization}
\begin{algorithmic}[1]
%\REQUIRE~ Affinity matrix $\textbf{W}$, maximum iteration $T$, error $\delta$
%\ENSURE~ Final discrete binary matching solution $\tilde{\textbf{x}}^*$
%\STATE \textbf{Input:} Matrix $\textbf{Z}\in \mathbb{R}^{n\times n}$, parameters $\gamma$,  maximum iteration $T_0$
%\STATE \textbf{Output:} Optimal mask $\textbf{M}\in \mathbb{R}^{n\times n}$
\STATE Initialize $\textbf{M} = \frac{1}{2\gamma}\textbf{Z}$
\WHILE{not convergence} %FOR {$t=1,2\cdots T'$}
\STATE
%$
%\widetilde{\textbf{M}} = \textbf{M}^{(t-1)} +\Big( \dfrac{1}{n}\textbf{I}+\dfrac{1}{n^2}\textbf{1} \textbf{M}^{(t-1)}
%\textbf{1}^{\mathrm{T}} -\dfrac{1}{n}\textbf{1}^{\mathrm{T}}\textbf{1}\textbf{M}^{(t-1)}  \Big)
%$
%\STATE Update ${\textbf{M}}^{(t)}$ as %\begin{center}
%$
%{\textbf{M}}^{(t)}_{ij}=\max\{0,\widetilde{\textbf{M}}_{ij}\}
%$
$
\,\,\,\,\,\,\,\,\,\,\,\,\, \widetilde{\textbf{M}} \leftarrow \textbf{M} -\dfrac{1}{n}\textbf{1}^{\mathrm{T}}\textbf{1} \textbf{M} + \textstyle  \big(\dfrac{1}{n}\textbf{I}+\dfrac{1}{n^2}\textbf{1} \textbf{M}
\textbf{1}^{\mathrm{T}} \textbf{I}-\dfrac{1}{n}\textbf{M}\big) \textbf{1}^{\mathrm{T}}\textbf{1}
$\\
 \STATE %Update ${\textbf{M}}$ as\\ %\begin{center}\\
$
\,\,\,\,\,\,\,\,\,\,\,\,\,{\textbf{M}}_{ij}\leftarrow \max \{\widetilde{\textbf{M}}_{ij},0\}
$
%\end{center}
\ENDWHILE %ENDFOR \label{code:recentEnd}
\STATE Discretization: Set ${\textbf{M}}_{ij}=1$ when $\textbf{M}_{ij}>\epsilon$, and ${\textbf{M}}_{ij}=0$ otherwise
\STATE Return
$
\textbf{M}
$
% {H}^{(k+1)} = \sigma\big( {F}^{(k)}  {W}^{(k)}\big)
%\STATE{Compute the final binary solution $\bar{\Xb}^*$ from $\Xb^*$ using a post-discretization step.}
\emph{}
\end{algorithmic}
\end{algorithm}

\textbf{Step 2}. Solving $\textbf{U}$ while fixing $\textbf{M}$, the problem becomes
\begin{align}\label{EQ:opt_U}
\min_{\textbf{U}} \,\, \mathrm{Tr} [\textbf{U}^{\mathrm{T}}(\textbf{I} - \textbf{M}\odot\hat{\textbf{A}}) \textbf{U}] + \mu\|\textbf{U}-\textbf{H}\|^2_F
\end{align}
This problem can be solved approximately via a $T$-step truncated power iteration algorithm (Eq.(6)) \cite{zhou2004learning} as follows,
\begin{align}\label{EQ:opt_U1}
{\textbf{U}}^{(t)} = \alpha(\textbf{M}\odot\hat{\textbf{A}}){\textbf{U}}^{(t-1)} + (1-\alpha)\textbf{H}, \,\,\, t=1,2\cdots T
\end{align}
where ${\textbf{U}}^{(0)}=\textbf{H}$ and $\alpha=\frac{1}{1+\mu}$.
One can also use a simple one-step iteration ($T=1$) to obtain an approximate solution efficiently as,
\begin{align}\label{EQ:opt_U10}
{\textbf{U}} = \alpha(\textbf{M}\odot\hat{\textbf{A}}){\textbf{H}} + (1-\alpha)\textbf{H}
\end{align}
which is similar to the feature aggregation used in GCN~\cite{kipf2016semi}. % with selective neighborhood. % mask weighting.
\section{GmCN Architecture}

\subsection{Layer-wise propagation}
The overall layer-wise propagation of GmCN is summarized in Algorithm 2, where $\sigma(\cdot)$ used in the last step denotes an activation function, such as $\mathrm{ReLU}(\cdot) = \max(0,\cdot)$.
Considering the efficiency of GmCN training, similar to~\cite{jiang2019graph} we employ a truncated iteration algorithm to optimize the $\textbf{M}$-problem approximately in GmCN architecture.
 GmCN can be used in many graph learning tasks. In this paper, we use it for semi-supervised classification.
%In this paper, we focus on semi-supervised classification.
% For this task, a softmax activation function is further used in the last layer to output the label prediction for graph nodes. %, defined as
Similar to many other works~\cite{kipf2016semi,velickovic2017graph}, the optimal weight matrix $\Theta$ of  each hidden layer in GmCN % $\{{\Theta}^{(0)},{\Theta}^{(1)},\cdots {\Theta}^{(K-1)}\}$
is trained by minimizing the cross-entropy loss via an Adam algorithm~\cite{Adam} which is initialized by using Glorot initialization~\cite{glorot2010understanding}.

\begin{algorithm}[h]
\caption{GmCN layer-wise propagation}
\begin{algorithmic}[1]
%\REQUIRE~ Affinity matrix $\textbf{W}$, maximum iteration $T$, error $\delta$
%\ENSURE~ Final discrete binary matching solution $\tilde{\textbf{x}}^*$
\STATE \textbf{Input:} Feature matrix $\textbf{H}\in \mathbb{R}^{n\times d}$, normalized graph $\hat{\textbf{A}}\in \mathbb{R}^{n\times n}$ and weight matrix $\Theta$,  parameter $\gamma,\alpha$ and $\epsilon$,  maximum iteration $K, L$ and $T$
\STATE \textbf{Output:} Feature map $\textbf{H}'$ % Optimal mask $\textbf{M}^*\in \mathbb{R}^{n\times n}$
\STATE Initialize $\textbf{U}= \textbf{H}$
%\STATE Compute $\textbf{M} = \frac{1}{2\gamma}\textbf{Z},\textbf{U}= \textbf{H}$
\FOR {$k=1,2\cdots K$}
% \STATE Compute $\textbf{Z} = \hat{\textbf{A}}\odot (\textbf{U}\textbf{U}^{\mathrm{T}})$
\STATE Compute $\textbf{M} = \frac{1}{2\gamma}\hat{\textbf{A}}\odot (\textbf{U}\textbf{U}^{\mathrm{T}})$

\FOR {$l=1,2\cdots L$}

\STATE Update ${\textbf{M}}$ as\\
$
\,\,\,\,\,\,\,\,\,\,\,\,\, \widetilde{\textbf{M}} \leftarrow \textbf{M} -\dfrac{1}{n}\textbf{1}^{\mathrm{T}}\textbf{1} \textbf{M} + \textstyle  \big(\dfrac{1}{n}\textbf{I}+\dfrac{1}{n^2}\textbf{1} \textbf{M}
\textbf{1}^{\mathrm{T}} \textbf{I}-\dfrac{1}{n}\textbf{M}\big) \textbf{1}^{\mathrm{T}}\textbf{1}
$\\
% \STATE Update ${\textbf{M}}$ as\\ %\begin{center}\\
$
\,\,\,\,\,\,\,\,\,\,\,\,\,{\textbf{M}}_{ij}\leftarrow \max \{0,\widetilde{\textbf{M}}_{ij}\}
$
%\end{center}
\ENDFOR  \label{code:recentEnd}
\STATE Discretize $\textbf{M}$ as: Set ${\textbf{M}}_{ij}=1$ when $\textbf{M}_{ij}>\epsilon$, and ${\textbf{M}}_{ij}=0$ otherwise
% \STATE Initialize $\textbf{U}= \textbf{H}$
\FOR {$t=1,2\cdots T$}

\STATE Update ${\textbf{U}}$ as\\
$
\,\,\,\,\,\,\,\,\,\,\,\,\,{\textbf{U}} \leftarrow \alpha({\textbf{M}}\odot\hat{\textbf{A}}){\textbf{U}} + (1-\alpha)\textbf{H}
$
%\end{center}
\ENDFOR \label{code:recentEnd}
\ENDFOR
\STATE Return
$
\textbf{H}' = \sigma(\textbf{U}\Theta) % {\textbf{M}}^{(T_m)}
$
% {H}^{(k+1)} = \sigma\big( {F}^{(k)}  {W}^{(k)}\big)
%\STATE{Compute the final binary solution $\bar{\Xb}^*$ from $\Xb^*$ using a post-discretization step.}
\emph{}
\end{algorithmic}
\end{algorithm}
\subsection{Complexity analysis}
The main computational complexity of GmCN involves mask $\textbf{M}$ optimization and
feature aggregation.
First, in $\textbf{M}$-optimization, each update has a simple solution which can be
computed efficiently.
In particular, since $\textbf{1}^{\mathrm{T}}\textbf{1}$ is a $n\times n$ matrix of ones,
thus the computational complexity is $\mathcal{O}(n^2)$ in the worst case (for the dense graph).
Second, we adopt a similar feature aggregation used in GCN~\cite{kipf2016semi} (as discussed in Section 3). Each update has the computational complexity as $\mathcal{O}(n^2d)$ in the worst case, where $d$ denotes the feature dimension of $\textbf{H}$.
Note that, in practical, both $\hat{\textbf{A}}$ and $\textbf{M}$ are usually sparse and thus the computations of these operations become more efficient.
Also, the maximum iterations $\{K, L, T\}$ are set to $\{4, 3, 3\}$ respectively and thus the overall propagation in GmCN is not computationally expensive.

\subsection{Comparison with related works}
% Here, we compare our GmCN with GCN~\cite{kipf2016semi}, GAT~\cite{velickovic2017graph}
First, in contrast to GCN~\cite{kipf2016semi}, the key idea behind GmCN is that it learns how to \emph{selectively} aggregate feature information from a
node’s local neighborhood set and thus conduct both
neighborhood learning and graph convolution simultaneously.
Second, our graph mask convolution (GmC) has some similarities with graph attention aggregation (GAT)~\cite{velickovic2017graph}.
The main differences are follows.
(1) GmC aims to select some neighbors for feature aggregation while GAT tries to assign different importances  to nodes
within the same neighborhood.
(2) We derive a unified regularization framework for the mask optimization and feature aggregation.
In contrast, in GAT, it employs  a single-layer feedforward neural network for graph attention optimization.
Third, neighborhood selection techniques in graph neural network training and learning have also been studied via some neighborhood sampling strategies~\cite{hamilton2017inductive,chen2017stochastic}. For example, in GraphSAGE~\cite{hamilton2017inductive}, it samples a fixed-size set of neighbors in feature aggregation.
Chen et al~\cite{chen2017stochastic} propose a more optimal
sampling technique with convergence analysis in GCN training.
Differently, in our GmCN, we derive neighborhood selection from a regularization framework and also conduct neighborhood selection and feature aggregation simultaneously in a unified manner.

\section{Experiments}

%In order to evaluate the effectiveness of the proposed GmCN model, we evaluate it on several widely used benchmark datasets and compare it with some other related methods. % at the same time, compare it with some other related methods.

\subsection{Datasets}

We evaluate GmCN on five widely used datasets including Cora~\cite{sen2008collective}, Citeseer~\cite{sen2008collective}, Cora-ML~\cite{mccallum2000automating},
Amazon Computers and Amazon Photo~\cite{mcauley2015image}.
In Cora, Citeseer and Cora-ML, nodes denote documents and edges encode the citation relationships between documents.
Amazon Computers and Amazon Photo are segments of the Amazon co-purchase graph~\cite{mcauley2015image}, in which nodes represent products and edges indicate that two products are frequently bought together. Each node is represented as bag-of-words encoded product reviews and its class label is given by the product category.
 The detail usages  of these datasets are introduced below.
% where nodes denote documents and edges encode the citation relationships between documents. And two Amazon networks datasets (Amazon Computers and Amazon Photo) are segments of the Amazon co-purchase graph~\cite{sen2008collective}), where nodes represent goods, and edges encode that two goods are frequently purchased together .
%
\begin{itemize}
  \item \textbf{Cora} contains 2708 nodes and 5429 edges. % in which nodes correspond to documents and edges encode the citation relationships among documents.
Each node is represented by a 1433 dimension feature descriptor and all the nodes are falling into six classes.
  \item
\textbf{Citeseer} contains 3327 nodes and 4732 edges.
Each node is represented by a 3703 dimension feature descriptor and all nodes are classified into six classes.
  \item
\textbf{Amazon Computers} contains 13381 nodes and 259159 edges.
Each node is represented by a 767 dimension feature descriptor and all the nodes are falling into ten classes.
  \item
\textbf{Amazon photo} contains 7487 nodes and 126530 edges. Each node is represented by a 745 dimension feature descriptor and
all the nodes are falling into eight classes.
  \item
\textbf{Cora-ML} contains 2810 nodes and 7981 edges.
Each node is represented by a 2879 dimension feature descriptor and all the nodes are falling into seven classes.
\end{itemize}

\subsection{Experimental setup}

% \subsubsection{Parameter setting}

% we follow the experimental setup of previous works ~\cite{kipf2016semi,velickovic2017graph}.
%\textbf{Parameter setting.} Similar to experimental setting of GCN~\cite{kipf2016semi,velickovic2017graph},
We use a two-layer graph convolutional network and the number of units in hidden layer is set to 16. Our GmCN is trained by using an ADAM algorithm~\cite{Adam}  with 10000 maximum epochs and learning rate of 0.01, as suggested in \cite{kipf2016semi}.
%We stop training if the validation loss does not decrease for 100 consecutive
%epochs, as suggested in \cite{kipf2016semi}.
% All the network weights are initialized using Glorot initialization~\cite{glorot2010understanding}.
The parameters $\{\alpha, \gamma\}$ are set to 0.8 and 0.001 respectively to obtain the best average performance.
%GmCN is not insensitive to these parameters. We provide additional experiments under different settings of parameters $\{\alpha$, $\gamma\}$  and network depth in \S 6.4.
The maximum iterations $\{K, L, T\}$ in Algorithm 2 are set to $\{4, 3, 3\}$, respectively.

% \subsubsection{Data setting}

\textbf{Data setting.}  For all datasets, we select 10\%, 20\% and 30\% nodes as labeled data and use the remaining nodes as unlabeled data.
For unlabeled data, we also use 10\% nodes for validation purpose to determine the convergence criterion, and use the remaining 80\%, 70\% and 60\% nodes respectively as testing samples.
All the reported results are averaged over five runs  with different data splits of training, validation and testing.
%
%we utilize the similar data setting used in previous works~\cite{kipf2016semi,velickovic2017graph}. For each class, we select 20 nodes as labeled data and 300 nodes as validation data, and then evaluate the performance of label prediction on the remaining 1000 test nodes.

\subsection{Comparison with state-of-the-art methods}

%\textbf{Baselines.}
%We first compare our GmCN with the baseline model GCN~\cite{kipf2016semi} to demonstrate the benefit of graph optimization.
%Also,
\textbf{Comparison methods}. We compare our GmCN method against some other recent graph neural network methods including DeepWalk~\cite{perozzi2014deepwalk}, Graph Convolutional Network (GCN)~\cite{kipf2016semi}, Graph Attention Networks (GAT)~\cite{velickovic2017graph}, CVD+PP~\cite{chen2017stochastic}, GraphSAGE~\cite{hamilton2017inductive}, APPNP~\cite{klicpera2018predict} and Deep Graph Informax (DGI)~\cite{DGI}.
For GraphSAGE, we use the mean aggregation rule.
Note that, both GraphSAGE~\cite{hamilton2017inductive} and CVD+PP~\cite{chen2017stochastic} also utilize neighborhood sampling techniques and thus are related with GmCN.
The codes of these methods were provided by authors and we use them in our experiments.

\textbf{Results}. Table 1-3 summarize the comparison results  on these benchmark datasets. % Table 2 and 3 summarize the comparison results on three image datasets. The best results are marked as bold.
Overall, one can note that, GmCN obtains the best results on all datasets.
In particular, we can note that
(1) GmCN outperforms GCN~\cite{kipf2016semi}, demonstrating the effectiveness of GmCN by incorporating the proposed neighborhood selection and optimization in GCN training. %, which indicates that GOCN conducts data representation and semi-supervised learning more optimal than GCN.
(2) GmCN performs better than GAT~\cite{velickovic2017graph} which utilizes an attention weighted feature aggregation in layer-wise propagation.
(3) GmCN performs better than other sampling GNNs including GraphSAGE~\cite{hamilton2017inductive} and CVD+PP~\cite{chen2017stochastic}.
This demonstrates the more effectiveness of the proposed neighborhood selection technique on selecting the useful  neighbors for GCN learning.
(4) GmCN outperforms recent graph network APPNP~\cite{klicpera2018predict} and DGI~\cite{DGI}, which demonstrates the advantages of GmCN on graph data representation and semi-supervised learning.
%
%recent graph network GAT ~\cite{velickovic2017graph} and DGI~\cite{DGI}, which demonstrates the benefit of GOCN on  data representation and learning.
%
%(2) GmCN performs better than recent graph network GAT ~\cite{velickovic2017graph} and DGI~\cite{DGI}, which demonstrates the benefit of GOCN on  data representation and learning.
%(3) GOCN generally performs better than other graph based semi-supervised method LP~\cite{zhu2003semi}, ManiReg~\cite{belkin2006manifold}
%and DeepWalk~\cite{perozzi2014deepwalk}, which further indicates the effectiveness of GOCN on conducting semi-supervised classification tasks.
%
\begin{table}[!htp]\small
	\centering
	\caption{\upshape Comparison results of different methods  on  Cora and Citeseer datasets. }
	\begin{tabular}{c|c|c|c|c|c|c}
		\hline
		\hline
		Dataset& \multicolumn{ 3}{c}{Cora} & \multicolumn{ 3}{|c}{Citeseer}\\
		\hline
		Ratio of label & 10\% & 20\% & 30\% & 10\% & 20\% & 30\% \\
		\hline
		Deep Walk& 74.83$\pm$0.45 & 78.58$\pm$1.22 & 79.34$\pm$1.45 & 45.26$\pm$2.21 & 48.16$\pm$1.93 & 48.75$\pm$0.89 \\
		APPNP       & 72.87$\pm$1.31 & 78.37$\pm$0.81 & 80.52$\pm$0.69 & 66.27$\pm$1.49 & 70.00$\pm$0.81 & 72.19$\pm$0.97 \\
		DGI      & 81.22$\pm$0.92 & 82.19$\pm$0.87 & 83.32$\pm$0.76 & 68.25$\pm$1.13 & 69.69$\pm$0.35 & 71.41$\pm$0.56 \\
		GraphSAGE      & 82.14$\pm$0.32 & 84.81$\pm$0.78 & 86.56$\pm$0.46 & 71.48$\pm$0.73 & 73.63$\pm$0.73 & 74.60$\pm$0.55 \\
		CVD+PP      & 80.21$\pm$0.88 & 84.03$\pm$0.77 & 85.48$\pm$0.45 & 71.91$\pm$0.74 & 73.29$\pm$0.99 & 74.71$\pm$0.31 \\
		GCN      & 80.23$\pm$0.66 & 84.28$\pm$0.50 & 85.53$\pm$0.40 & 71.49$\pm$1.00 & 72.99$\pm$0.77 & 74.31$\pm$0.35 \\
		GAT      & 82.51$\pm$1.15 & 84.42$\pm$0.85 & 85.70$\pm$0.83 & 69.08$\pm$0.84 & 69.61$\pm$0.84 & 72.64$\pm$0.44 \\
		\hline
		GmCN     & \textbf{83.07$\pm$0.56} & \textbf{85.70$\pm$0.56} & \textbf{87.29$\pm$0.42} & \textbf{72.30$\pm$0.37} & \textbf{74.13$\pm$0.74} & \textbf{74.92$\pm$0.34} \\
		\hline
		\hline
	\end{tabular}
\end{table}
\begin{table}[!htp]\small
	\centering
	\caption{\upshape Comparison results of methods  on  Amazon Computers and Amazon Photo datasets. }
	\begin{tabular}{c|c|c|c|c|c|c}
		\hline
		\hline
		Dataset& \multicolumn{ 3}{c }{Amazon Computers} & \multicolumn{ 3}{|c}{Amazon Photo}\\
		\hline
		Ratio of label & 10\% & 20\% & 30\% & 10\% & 20\% & 30\% \\
		\hline
		Deep Walk& 87.01$\pm$0.53 & 87.91$\pm$0.13 & 88.31$\pm$0.15 & 90.47$\pm$0.38 & 91.63$\pm$0.43 & 91.84$\pm$0.14 \\
		APPNP       & 85.52$\pm$0.67 & 86.99$\pm$0.56 & 87.21$\pm$0.53 & 91.57$\pm$0.44 & 92.91$\pm$0.25 & 92.97$\pm$0.49\\
		DGI      & 85.32$\pm$0.47 & 86.31$\pm$0.28 & 86.30$\pm$0.42 & 92.53$\pm$0.24 & 92.80$\pm$0.30 & 92.64$\pm$0.23 \\
		GraphSAGE      & 82.75$\pm$0.57 & 83.88$\pm$0.97 & 84.75$\pm$0.41 & 89.78$\pm$0.83 & 91.51$\pm$0.62 & 91.96$\pm$0.34 \\
		CVD+PP      & 83.71$\pm$1.53 & 85.72$\pm$0.43 & 81.14$\pm$1.12 & 91.52$\pm$0.40 & 92.53$\pm$0.38 & 92.27$\pm$0.28 \\
		GCN      & 85.51$\pm$0.49 & 85.28$\pm$1.29 & 84.67$\pm$0.76 & 92.58$\pm$0.25 & 92.65$\pm$0.44 & 92.64$\pm$0.37 \\
		GAT      & 83.64$\pm$5.21 & 86.10$\pm$2.50 & 85.70$\pm$3.05 & 90.53$\pm$2.91 & 91.26$\pm$1.14 & 91.93$\pm$1.11 \\
		\hline
		GmCN     & \textbf{89.14$\pm$0.70} & \textbf{89.43$\pm$0.74} & \textbf{89.88$\pm$0.52} & \textbf{93.80$\pm$0.18} & \textbf{94.20$\pm$0.21} & \textbf{94.31$\pm$0.28} \\
		\hline		
		\hline
	\end{tabular}
\end{table}
\begin{table}[!htp]\small
	\centering
	\caption{\upshape Comparison results of different methods  on  Cora-ML dataset.}
	\begin{tabular}{c|c|c|c}
		\hline
		\hline
		Dataset& \multicolumn{ 3}{c}{Cora-ML} \\
		\hline
		Ratio of label & 10\% & 20\% & 30\% \\
		\hline
		Deep Walk& 79.13$\pm$1.60 & 81.51$\pm$0.89 & 82.36$\pm$0.90  \\
		APPNP     & 76.96$\pm$0.46 & 80.02$\pm$0.93 &81.43$\pm$0.42 \\
		DGI      & 84.45$\pm$0.49 & 86.23$\pm$0.62 & 86.74$\pm$0.45  \\
		GraphSAGE     & 84.75$\pm$0.76 & 86.62$\pm$0.55 & 87.67$\pm$0.81  \\
		CVD+PP     & 84.97$\pm$0.40 & 87.00$\pm$0.40 & 87.25$\pm$0.74  \\
		GCN      & 85.17$\pm$0.70 & 86.92$\pm$0.32 & 87.23$\pm$0.77  \\
		GAT      & 81.53$\pm$1.78 & 83.46$\pm$0.99 & 83.89$\pm$0.57  \\
		\hline
		GmCN     & \textbf{86.02$\pm$0.48} & \textbf{87.41$\pm$0.65} & \textbf{88.05$\pm$0.81} \\
		\hline				
		\hline
	\end{tabular}
\end{table}

\subsection{Parameter analysis}

% In this section, we evaluate the performance of GmCN model with different settings of network parameters. %under different network settings and parameters.
%
% We first investigate the influence of GmCN model depth (number of convolutional layers).
Figure 1 (a) shows the performance of GmCN model across different number of hidden layers on Cora dataset.
% As a baseline, we also report the results of GCN model with the same depth setting.
Note that, GmCN maintains better performance under different model depths and consistently performs better than GCN, which indicates the insensitivity of the GmCN w.r.t. model depth.
%
% We then investigate the influence of the network parameter $\alpha$ and $\gamma$. % graph learning parameter $\lambda$ in our GLCN.
Table 4 and 5 show the performance of GmCN with different parameter settings of parameter $\alpha$ and $\gamma$, respectively,  which
indicate the insensitivity of the GmCN w.r.t. parameter $\alpha$ and $\gamma$.
\begin{table}[!htp]
	\centering
	\caption{\upshape Results of GmCN across different parameter $\alpha$ values on Cora and Citeseer datasets. }
	\begin{tabular}{c|c|c|c|c}
		\hline
		\hline
		$\alpha$ & 0.6 & 0.7 & 0.8 & 0.9 \\
		\hline
		Cora   & 78.45$\pm$0.93 & 80.06$\pm$1.01 & 83.07$\pm$0.56 & 82.66$\pm$0.72 \\
		\hline
		Citeseer   & 72.45$\pm$0.33 & 73.46$\pm$0.53 & 72.52$\pm$0.63 & 71.12$\pm$0.44 \\
		\hline				
		\hline
	\end{tabular}
\end{table}
\begin{table*}
	\centering
	\caption{\upshape  Results of GmCN across different parameter $\gamma$ values on Cora and Citeseer datasets. }
	\begin{tabular}{c|c|c|c|c}
		\hline
		\hline
		$\gamma$ & 0.0001 & 0.0005 & 0.001 & 0.005 \\
		\hline
		Cora   & 83.21$\pm$0.81 & 83.34$\pm$0.68 & 83.13$\pm$0.80 & 83.00$\pm$0.97 \\
		\hline
		Citeseer   & 72.09$\pm$0.42 & 72.50$\pm$0.26 & 72.76$\pm$0.25 & 72.85$\pm$0.54 \\
		\hline				
		\hline
	\end{tabular}
\end{table*}
\begin{figure}[!htp]
\centering
%\subfigure[lambda]{
%\begin{minipage}{6cm}
\centering
\subfigure[Results across different number of hidden layers]{\includegraphics[width=0.48\textwidth]{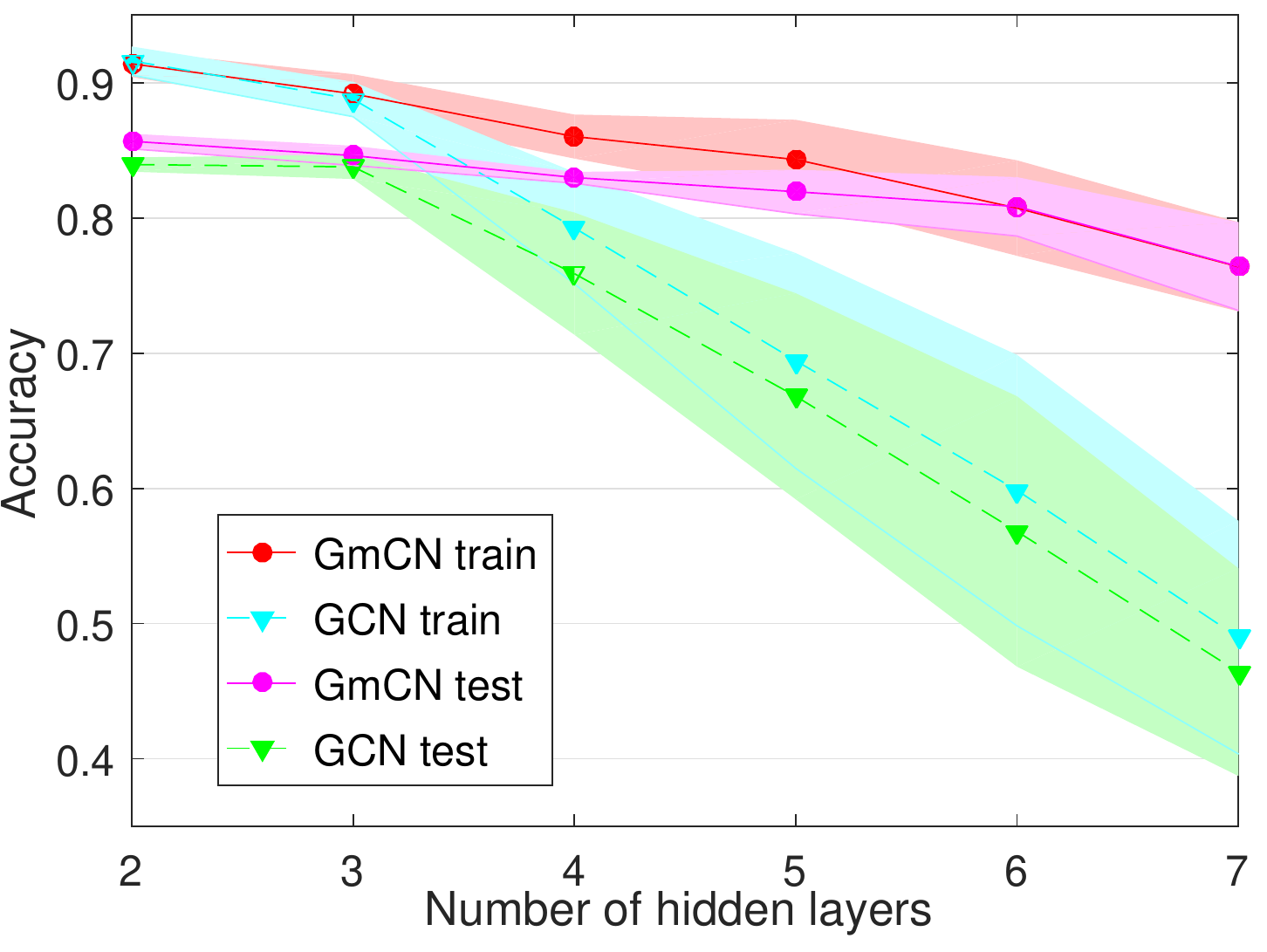}}\subfigure[Results on different levels of graph noise]{\includegraphics[width=0.48\textwidth]{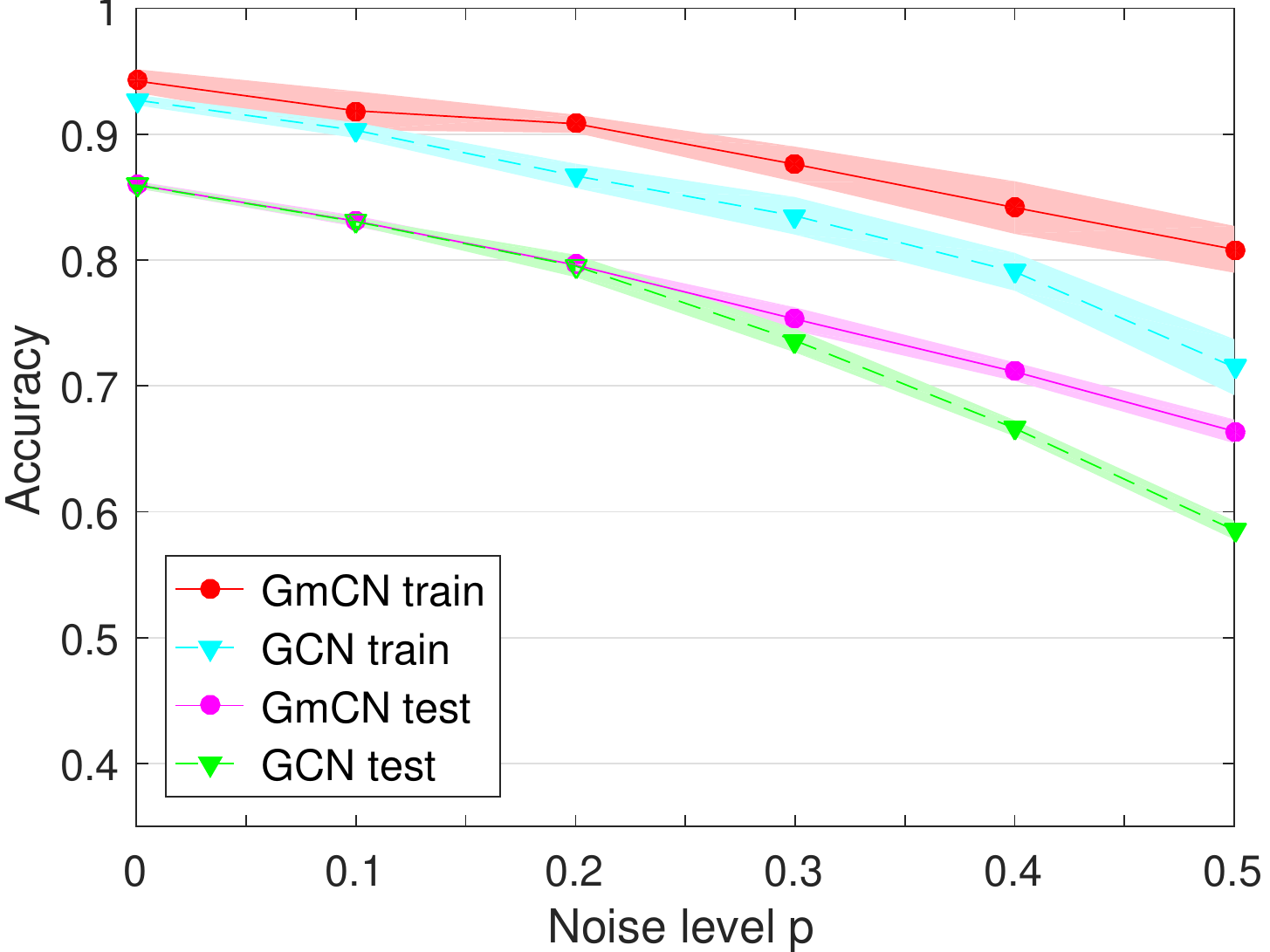}}
% }
  \caption{Results of GmCN on different parameters and graph noise levels on Cora dataset.}\label{fig::lambda}
\end{figure}

\subsection{Perturbation analysis}

To evaluate the robustness of the proposed GmCN w.r.t. graph noises,
we test our method on graph data with different perturbation noise levels. The perturbation is built as follows. We first randomly cut each edge and then connect
each pair of randomly selected nodes  with probability $p$. Thus,
parameter $p$ controls the level of structure perturbation.
For each level $p$, we generate ten noisy graphs and
then compute the average performances including both training and testing accuracy.
 As a baseline, we also list the results of GCN model on the same noise data.
Figure 1 (b) shows the comparison results on Cora dataset.
We can note that GmCN performs more robustly on noisy graph, which further demonstrates  that
the proposed mask selection  mechanism is helpful for capturing important information and
thus avoiding noisy graph information.

\section{Conclusion}

This paper proposes a novel Graph mask Convolutional Network (GmCN) for graph data representation and semi-supervised learning.
GmCN learns an optimal graph neighborhood structure that better serves GCN based semi-supervised learning.
A new unified regularization framework is derived for GmCN interpretation, based on which we provide an effective update algorithm
to achieve GmCN optimization.
%
%GOCN is inspired based on the re-formulation of graph convolution as a regularization framework.
%% GLCN aims to learn an optimal graph structure that best serves graph CNNs for semi-supervised learning
%GOCN integrates graph optimization and convolution in a unified scheme and thus can boost their respectively performance in graph neural network learning.
%% the proposed new graph learning operation and traditional graph convolution architecture together in a unified network,
%%
%%which can learn an optimal graph structure that best serves GCN for semi-supervised learning problem.
%%
%Experimental results on several benchmarks demonstrate
%the effectiveness and robustness of  GmCN  on semi-supervised learning tasks.
%In the future, we will explore GmCN on some other unsupervised learning tasks, such as clustering, embedding etc.

% Therefore, the overall complexity is less than $T\mathcal{O}(n^2d)+$
\bibliographystyle{ieee}
\bibliography{nmfgm-SGCN}

\end{document}